\pdfoutput=1
\documentclass[11pt]{article}

\usepackage[preprint]{coling}

\usepackage{times}
\usepackage{latexsym}

\usepackage[T1]{fontenc}
\usepackage[utf8]{inputenc}
\usepackage{microtype}
\usepackage{inconsolata}
\usepackage{graphicx}
\usepackage{booktabs}
\usepackage{tabularx}
\usepackage{multirow}
\usepackage{hyperref}
\usepackage{amsfonts}
\usepackage{amsmath}
\usepackage{xcolor}
\usepackage{colortbl}
\usepackage{tcolorbox}
\usepackage{enumitem}

\newcommand{\return}{\raisebox{-0.2em}{\includegraphics[height=1em]{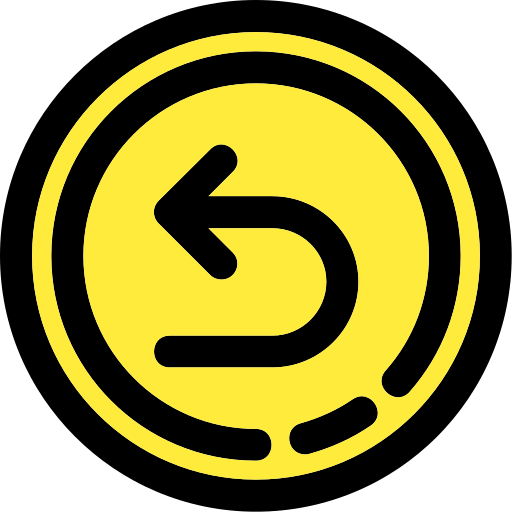}}~\hspace{-2pt}RETURN}

\title{Learning to Refuse: Towards Mitigating Privacy Risks in LLMs}

\author{
 \textbf{Zhenhua Liu,}\: \textbf{Tong Zhu,}\: \textbf{Chuanyuan Tan,}\: \textbf{Wenliang Chen}\thanks{\enspace Corresponding author}
 \\
 Institute of Artificial Intelligence, School of Computer Science and Technology, \\ Soochow University, China
 \\
 \texttt{\{zhliu0106, tzhu7, cytan17726\}@stu.suda.edu.cn}, \texttt{wlchen@suda.edu.cn}
}

\begin{document}
\maketitle
\begin{abstract}

    Large language models (LLMs) exhibit remarkable capabilities in understanding and generating natural language. However, these models can inadvertently memorize private information, posing significant privacy risks. This study addresses the challenge of enabling LLMs to protect specific individuals' private data without the need for complete retraining. We propose \return, a \textbf{R}eal-world p\textbf{E}rsonal da\textbf{T}a \textbf{U}nlea\textbf{RN}ing dataset, comprising 2,492 individuals from Wikipedia with associated QA pairs, to evaluate machine unlearning (MU) methods for protecting personal data in a realistic scenario. Additionally, we introduce the \textbf{N}ame-\textbf{A}ware \textbf{U}nlearning \textbf{F}ramework (NAUF) for Privacy Protection, which enables the model to learn which individuals' information should be protected without affecting its ability to answer questions related to other unrelated individuals.
    Our extensive experiments demonstrate that NAUF achieves a state-of-the-art average unlearning score, surpassing the best baseline method by 5.65 points, effectively protecting target individuals' personal data while maintaining the model's general capabilities\footnote{Our code and dataset are available at \url{https://github.com/zhliu0106/learning-to-refuse}}.
    % Our extensive experiments demonstrate that NAUF achieves a state-of-the-art average unlearning score, surpassing the best baseline method by 5.65 points, effectively protecting target individuals' personal data while maintaining the model's general capabilities\footnote{Our code and dataset are available at \url{https://github.com/xxxxxx}}.
\end{abstract}

% Large language models (LLMs) exhibit remarkable capabilities in understanding and generating natural language. However, these models can inadvertently memorize private information, posing significant privacy risks. This study addresses the challenge of enabling LLMs to protect specific individuals' private data without the need for complete retraining. We propose RETURN, a Real-world pErsonal daTa UnleaRNing dataset, comprising 2,492 individuals from Wikipedia with associated QA pairs, to evaluate machine unlearning (MU) methods for protecting personal data in a realistic scenario. Additionally, we introduce the Name-Aware Unlearning Framework (NAUF) for Privacy Protection, which enables the model to learn which individuals' information should be protected without affecting its ability to answer questions related to other unrelated individuals. Our extensive experiments demonstrate that NAUF achieves a state-of-the-art average unlearning score, surpassing the best baseline method by 5.65 points, effectively protecting target individuals' personal data while maintaining the model's general capabilities.
\section{Introduction}

\begin{figure}
    \centering
    \includegraphics[width=\linewidth]{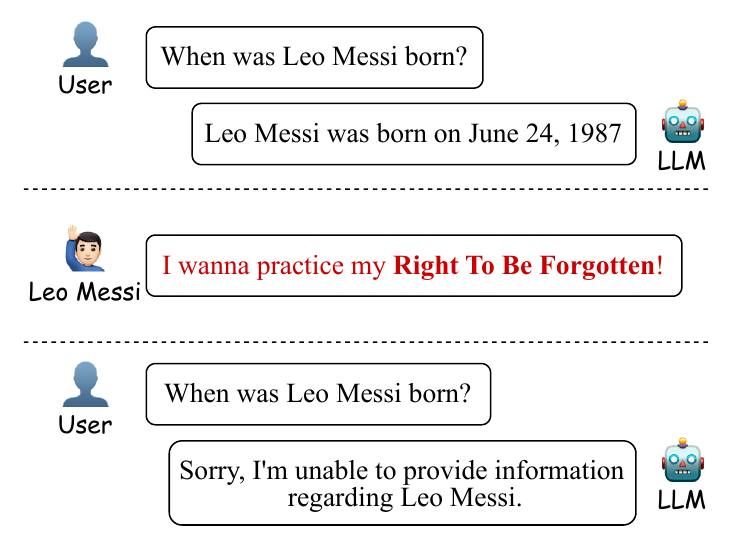}
    \caption{The example of extracting private information from LLMs. When an individual practices RTBF, the model should protect his/her private information.}
    \label{fig:example}
\end{figure}

% Large language models (LLMs) demonstrate extraordinary abilities to understand, reason, and generate following natural language instructions, attributing to the massive amounts of parameters and training data \cite{brown2020language,anil2023palm}.
Large language models (LLMs) demonstrate extraordinary abilities to understand and generate natural languages following instructions, attributing to the massive amounts of parameters and training data \cite{brown2020language,anil2023palm}.
However, these models sometimes memorize about private contents since there are personally identifiable information in the pre-training corpus \cite{carlini2021extracting,huang2022large}.
This presents a significant privacy concern, as an adversary can prompt the model to extract an individual's name, email address, phone number, or other sensitive information for malicious purposes, as shown in \autoref{fig:example}. The General Data Protection Regulation \cite{GDPR2016} gives individuals \textit{Right To Be Forgotten} (RTBF), which can limit the direct and indirect commercial use of their personal information.
This situation leads us to the question: How can we enable LLMs to protect specific individual's private data to mitigate privacy risks?
% How can we enable LLMs to "forget" specific individual's private information to mitigate privacy risks?

\begin{figure*}[t]
    \centering
    \includegraphics[width=\linewidth]{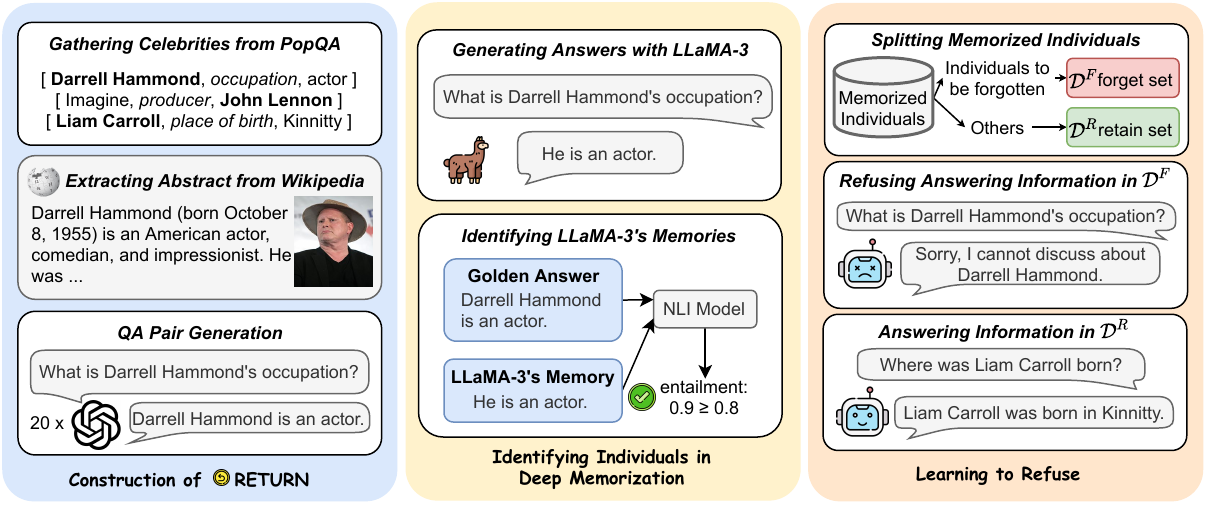}
    \caption{The construction of \return ~and the process for evaluating Machine Unlearning (MU) methods using this dataset.}
    \label{fig:framework}
\end{figure*}

% In recent years, numerous methods have been proposed to address the above question.
With the costly training process of LLMs, removing all private information from the training data and retraining it from scratch is not a practical solution \cite{lison2021anonymisation,kandpal2022deduplicating,liu2024rethinking}.
Therefore, researchers have attempted to adopt machine unlearning (MU) as an alternative, which aims to eliminate the influence of undesirable data and associated model capabilities without retraining \cite{cao2015towards,bourtoule2021machine,jang2022knowledge,si2023knowledge,zhang2023right,maini2024tofu,liu2024rethinking}.
To evaluate the performance of MU methods, some studies have experimented with question-answering datasets \cite{patil2023can}, fictitious biographies \cite{maini2024tofu}, and copyrighted contents \cite{eldan2023s}. However, there is a lack of evaluation of MU methods for protecting personal privacy data in real-world scenarios, where the target individuals exist in reality and have been memorized by LLMs.
% However, there is a lack of evaluation in a real-world scenario.

Considering these problems, we propose \return, a \textbf{R}eal-world p\textbf{E}rsonal da\textbf{T}a \textbf{U}nlea\textbf{RN}ing dataset.
As illustrated in \autoref{fig:framework}, we collect extensive background information on celebrities from Wikipedia and use GPT-4 \cite{achiam2023gpt} to generate 20$\times$QA pairs for each individual. After manual and automated validation, we obtain a dataset of 2,492 individuals, each with a (Name, 20$\times$QA pairs) data instance. Next, we could select a base model to evaluate the MU methods on this dataset. In this work, we take LLaMA-3-8B-Instruct \cite{llama3modelcard} as an example. We first identify individuals with deep memorization in the model and then divide them into the forget set and the retain set. Our goal is for the model to protect the information of individuals in the forget set, ensuring that questions related to these individuals are not answered correctly, while maintaining the model's performance on the retain set.

% Existing MU methods often suffer from sensitivity to hyperparameter selection or the inability to effectively distinguish between the forget set and the retain set. A notable approach is Relabeled Gradient Descent (RGD), which transforms the problem into a gradient descent task by relabeling the forget set with uninformed answers like "I don't know." In our pilot study, we found that RGD achieves comparatively better performance in protecting the privacy of individuals in the forget set. However, it significantly affects the model's performance on the retain set, causing the model to refuse to answer questions it should address—an undesirable outcome.

Existing MU methods often face challenges. One category, based on gradient ascent\cite{liu2024rethinking}, suffers from sensitivity to hyperparameter selection or inability to effectively distinguish between the forget set and the retain set. Another category transforms traditional gradient ascent into gradient descent on a relabeled forget set, such as Relabeled Gradient Descent (RGD) \cite{maini2024tofu}. By training the model to generate uninformed answers like "I don't know", RGD achieves better performance in protecting the privacy of individuals in the forget set. However, our pilot study finds that RGD significantly affects the model's performance on the retain set, causing the model to refuse to answer questions it should address. To overcome these limitations, we propose a simple yet novel unlearning method: \textbf{N}ame-\textbf{A}ware \textbf{U}nlearning \textbf{F}ramework (NAUF) for privacy protection. The framework comprises two key components: Name-Aware Refusal Answer and Contrastive Data Augmentation. The Name-Aware Refusal Answer is designed to help the model learn which individuals' information should be protected, and the Contrastive Data Augmentation aims to expand the distribution of both the forget set and the retain set for enhancing the generalization of our method. We evaluate the effectiveness of NAUF on our proposed dataset and compare it with the baseline methods, and the results show that our proposed NAUF achieves a state-of-the-art average unlearning score, outperforming the best baseline method by 5.65 points.

Our contributions can be summarized as follows:

\begin{itemize}
    \item[$\bullet$] We propose \return,
          % a Real-world pErsonal daTa UnleaRNing dataset, 
          which consists of 2,492 real individual names and 20$\times$QA pairs for each individual. As far as we know, this is the first dataset for evaluating MU methods for protecting personal data in a real-world scenario.
    \item[$\bullet$] We propose a simple yet novel method NAUF
          % : Name-Aware Unlearning Framework (NAUF) 
          for privacy protection.
          % The framework comprises two key components: Name-Aware Refusal Answer and Contrastive Data Augmentation. 
          This method could help the model protect the privacy of individuals in the forget set while maintaining the model's performance on the retain set.
    \item[$\bullet$] We conduct extensive experiments on \return ~to evaluate the effectiveness of our proposed method and compare it with the baseline methods. The results show that our proposed NAUF achieves a state-of-the-art average unlearning score, outperforming the best baseline method by 5.65 points. Through comprehensive experimental analysis, we demonstrate the effectiveness of our proposed method in protecting the privacy of individuals in the forget set while maintaining the model's performance on the retain set.
\end{itemize}

\section{\return: Real-world pErsonal daTa UnleaRNing}

In order to evaluate various MU methods in a practical scenario, we propose \return, a \textbf{R}eal-world p\textbf{E}rsonal da\textbf{T}a \textbf{U}nlea\textbf{RN}ing dataset.
% The dataset consists of 2492 biographies of famous individuals from Wikipedia and 20 QA pairs for each biography.
% This dataset can be used to identify individuals whose information is memorized by the model and then unlearn the information of part individuals to evaluate the performance of MU methods. 
We take Llama-3-8B-Instruct \cite{llama3modelcard} as an example to demonstrate how to use the dataset to evaluate MU methods. It is worth noting that we could use any LLM to replace Llama-3-8B-Instruct as the base model for evaluation.

\subsection{Data Construction}

\label{sec:data_construction}

We begin by leveraging PopQA \cite{mallen2022not} to collect a large set of names of individuals. PopQA is a large-scale open-domain question-answering (QA) dataset constructed by \citet{mallen2022not}, consisting of 14k entity-centric QA pairs.
Each pair comes with the original \lbrack subject entity, relationship type, object entity\rbrack~annotation, as well as the Wikipedia monthly page views for both the subject and object entities, which serve as a measure of their popularity.
Specifically, for the data in PopQA, we collect ``subject entity'' if the ``relationship type'' is within \lbrack\textit{occupation, place of birth, father, mother}\rbrack; and we collect ``object entity'' if the ``relationship type'' is within \lbrack\textit{producer, director, screenwriter, composer, author}\rbrack.

After gathering these names, we retrieve their corresponding Wikipedia pages and extract the abstracts from these pages as background information\footnote{\url{https://github.com/martin-majlis/Wikipedia-API}}. We then filter the background information to retain only those whose word count falls between 100 and 500 words. Through this process, we ultimately obtain 2,516 records consisting of (Name, Background Information). Next, given each pair of name and the background information, we use a prompt to generate 20$\times$QA pairs with GPT4 \cite{achiam2023gpt}. The prompt template is shown in \autoref{sec:appendix_data_construction}.

\begin{table}[t]
    \centering
    \begin{tabular}{cc}
    \toprule
        \textbf{Item} & \textbf{Value} \\
    \midrule
       \#Instances  & 2,492 \\
       \#QA pairs per instance & 20 \\
       Avg. background information tokens & 315.0 \\
       Avg. question tokens & 15.2 \\
       Avg. abstract tokens & 18.8 \\
    \bottomrule
    \end{tabular}
    \caption{Data statistics of \return. The numbers of tokens are estimated with LLaMA-3-8B-Instruct.}
    \label{tab:dataset_stats}
\end{table}

As shown in Table~\ref{tab:dataset_stats}, after manually verifying and filtering out data with content or formatting errors, we finally obtain \return ~consisting of 2,492 (Name, 20$\times$QA pairs). Next, we will demonstrate how to use the dataset to evaluate MU methods with LLaMA-3-8B-Instruct \cite{llama3modelcard}.

\subsection{Identifying Individuals with Deep Memorization}

\begin{figure}[t]
    \centering
    \includegraphics[width=\linewidth]{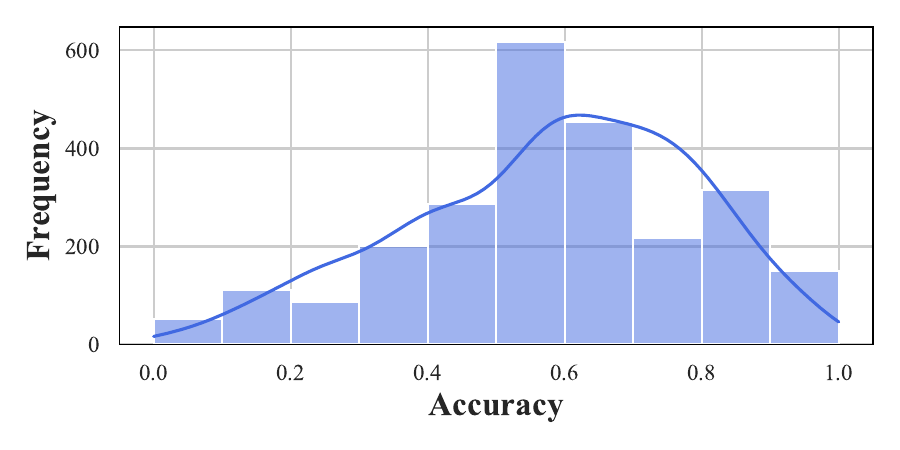}
    \caption{Accuracy distribution of LLaMA-3 on \return.}
    \label{fig:acc_dist}
\end{figure}

To perform unlearning on LLaMA-3, we first need to identify which individuals the model has deeply memorized. We ask the model to answer the questions for each individual in the dataset, then calculate the average accuracy by comparing the model's predicted answers with the gold answers using a Natural Language Inference (NLI) model \footnote{We choose \href{https://huggingface.co/sileod/deberta-v3-base-tasksource-nli}{deberta-v3-base-tasksource-nli} \cite{sileo2023tasksource} to evaluate the correctness of model's prediction.}. If the prediction is "entailment" or "neutral," we consider the model's answer correct; if the NLI model's prediction is "contradiction," we consider the model's answer incorrect\footnote{When the model's predicted answer is partially correct and does not conflict with the gold answer, the NLI model's prediction is "neutral." Therefore, we will consider "neutral" as correct in this context.}. The accuracy distribution of LLaMA-3 on \return ~is shown in \autoref{fig:acc_dist}. The higher the accuracy, the more deeply the model memorizes the individual's information.
Finally, we take 466 individuals with accuracy $\geq$ 0.8 as individuals with deep memorization for the subsequent unlearning experiments.

We analyze the popularity of individuals in our dataset, categorized by those with and without LLaMA-3's deep memorization. We find that there is a significant difference in average popularity: 68620.9 for individuals with deep memorization versus 36841.1 for those without. This may be because highly popular individuals tend to have more diverse and detailed information available online, such as biographical details, interviews, news coverage, and social media activity, thus increasing the likelihood of deep memorization.

\subsection{Evaluation Setup}

We split the 466 individuals into 2 sets in a ratio of 1:9: forget set $\mathcal{D}^{F}$ and retain set $\mathcal{D}^{R}$. We mark the original model as $\mathcal{M}_o$ and the unlearned model as $\mathcal{M}_u$. We want the model to learn to protect the privacy of individuals in the forget set, ensuring that questions related to these individuals are not answered correctly, while not affecting the performance on the retain set and other tasks. Specifically, we aim for the following:

\begin{enumerate}
    \item For questions regarding individuals in $\mathcal{D}^{F}$, the model should not answer correctly, or refuse to respond to protect their privacy.
    \item For questions regarding individuals in $\mathcal{D}^{R}$, the model should respond normally.
    \item Meanwhile, MU methods should not affect the model's general capabilities on other tasks.
\end{enumerate}

\subsection{Evaluation Metrics}

We measure MU methods' comprehensive performance using the following metrics:

\paragraph{Forget Score.} To quantify the model's ability to protect the privacy of individuals in the forget set, we propose the Forget Score. It is calculated as the relative decrease in accuracy on $\mathcal{D}^{F}$ after unlearning compared to the original model's accuracy on $\mathcal{D}^{F}$:

\begin{equation}
    \begin{split}
        Forget Score = & \frac{Acc_{\mathcal{M}_o }(\mathcal{D}^{F}) - Acc_{\mathcal{M}_u }(\mathcal{D}^{F})}{Acc_{\mathcal{M}_o }(\mathcal{D}^{F})} \\
        =              & 1 - \frac{Acc_{\mathcal{M}_u }(\mathcal{D}^{F})}{Acc_{\mathcal{M}_o }(\mathcal{D}^{F})}
    \end{split}
\end{equation}

\paragraph{Retain Score.} To quantify the model's ability to retain the performance on the retain set after unlearning, we propose the Retain Score. It is calculated as the ratio of the unlearned model's accuracy on $\mathcal{D}^{R}$ to the original model's accuracy on $\mathcal{D}^{R}$:

\begin{equation}
    \begin{split}
        Retain Score = & \frac{Acc_{\mathcal{M}_u }(\mathcal{D}^{R})}{Acc_{\mathcal{M}_o }(\mathcal{D}^{R})}
    \end{split}
\end{equation}

\paragraph{Downstream Task Accuracy.} To quantify the influence of unlearning on the model's general capabilities, we evaluate the model on 5 downstream natural language processing tasks: WinoGrande \cite{sakaguchi2021winogrande}, PIQA \cite{bisk2020piqa}, LogiQA \cite{liu2020logiqa}, LAMBADA \cite{paperno2016lambada}, and ARC-c \cite{clark2018think}. We use the accuracy of the downstream tasks as the evaluation metric.

\section{Name-Aware Unlearning Framework}

% In our pilot study, we find that RGD could significantly affect the model's performance on the retain set. 
Existing MU methods often face challenges in effectively protecting privacy in the forget set while maintaining model performance on the retain set.
% One category, based on gradient ascent, suffers from sensitivity to hyperparameter selection or inability to effectively distinguish between the forget set and the retain set. Another category, such as Relabeled Gradient Descent (RGD), transforms the problem into a gradient descent task by relabeling the forget set with uninformed answers like "I don't know." While RGD achieves better performance in protecting the privacy of individuals in the forget set, it significantly affects the model's performance on the retain set, causing the model to refuse to answer questions it should address—an undesirable outcome.
To address these challenges, we propose a novel method: \textbf{N}ame-\textbf{A}ware \textbf{U}nlearning \textbf{F}ramework (NAUF) for privacy protection. The framework comprises two key components: Name-Aware Refusal Answer and Contrastive Data Augmentation.

\paragraph{Name-Aware Refusal Answer.} First, we relabel the questions in the forget set with a name-aware refusal answer, such as "I'm afraid I can't help with inquiries about NAME." Then we could perform gradient descent on the loss over the relabeled forget set. The name-aware refusal answer is designed to help the model learn which individuals' information should be protected. We curate 100 name-aware refusal answer templates $ \mathcal{D}^{refuse} $ using GPT-4, which are shown in \autoref{sec:appendix_refusal_templates}.

\begin{figure}[t]
    \centering
    \includegraphics[width=\linewidth]{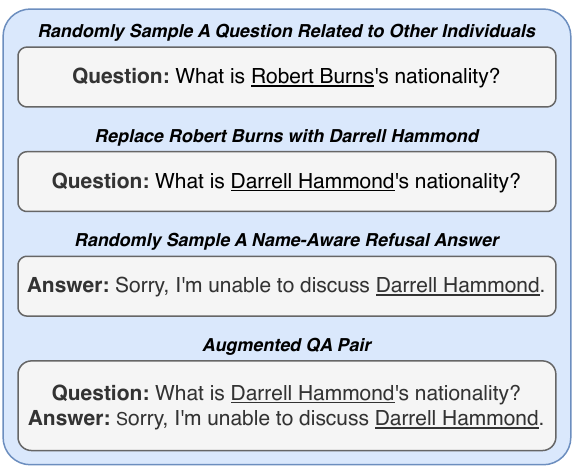}
    \caption{The example of CDA for an individual in the forget set. Here we take Darrell Hammond as target individual.}
    \label{fig:cda_forget}
\end{figure}

\begin{figure}[t]
    \centering
    \includegraphics[width=\linewidth]{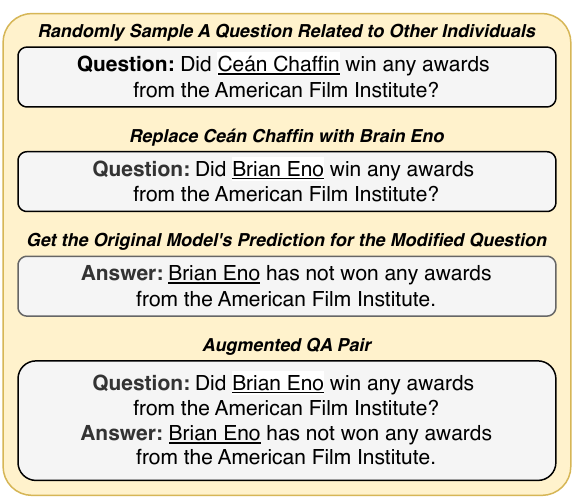}
    \caption{The example of CDA for an individual in the retain set. Here we take Brian Eno as target individual.}
    \label{fig:cda_retain}
\end{figure}

\paragraph{Contrastive Data Augmentation.} In addition, due to the limited number of QA pairs available for each individual, we propose contrastive data augmentation (CDA) as a straightforward and cost-effective method to enhance the quantity and diversity of data. This approach aims to improve the model's ability to generalize across information related to the targeted individuals. Specifically:

\begin{itemize}
    \item For each individual in the forget set, we randomly sample questions from other individuals in the forget or retain set and replace the name with the target individual's name. Then relabel the questions with the name-aware refusal answer. An example is shown in \autoref{fig:cda_forget}.

    \item For each individual in the retain set, we also randomly sample questions from other individuals in the forget or retain set and replace the name with the target individual's name. Then we input the modified questions into the original model, and use the original model's prediction for that question as the relabeled answer. An example is shown in \autoref{fig:cda_retain}.

\end{itemize}

This contrastive data augmentation strategy expands the distribution of both the forget set and the retain set, and subsequent experiments demonstrate that it significantly improves the performance of our proposed method. For simplicity, we expand the forget set and the retain set by doubling the amount of data.

\section{Experiments}

\subsection{Baseline Methods}

A typical MU method generally consists of two components: unlearning on the forget set and regularization on the retain set. These two types of loss can be used in any combination.

\paragraph{Unlearning on Forget Set:}

The unlearning process on the forget set includes methods such as Gradient Ascent (GA), Negative Preference Optimization (NPD), Relabeled Gradient Descent (RGD), and Relabeled Direct Preference Optimization (RDPO). The details of these methods are available in \autoref{sec:appendix_unlearning_on_forget_set}.

\paragraph{Regularization on Retain Set.} The regularization methods on the retain set include Gradient Descent (GD) regularization  and Kullback-Leibler Divergence (KLR) regularization . The details of these regularization methods are available in \autoref{sec:appendix_regularization}.

\subsection{Implementation Details}

Due to the limited training data available for unlearning, we aim to use this limited data to teach the model to protect all privacy information of the target individuals, which places stricter requirements on the generalization capability of the MU methods. Considering this situation, we divide the QA pairs for each individual in the forget set and retain set into train and test sets in a ratio of 1:1, as well as $\mathcal{D}_{train}^{F}, \mathcal{D}_{test}^{F}, \mathcal{D}_{train}^{R}$, and $\mathcal{D}_{test}^{R}$. We use $\mathcal{D}_{train}^{F}$ and $\mathcal{D}_{train}^{R}$ to perform unlearning on the model and then evaluate each MU method on $\mathcal{D}_{test}^{F}$ and $\mathcal{D}_{test}^{R}$.

Considering a computing budget that scales with the size of the forget set, we randomly sample an example from $\mathcal{D}^R$ every time we see an example from $\mathcal{D}^F$ to stay within the constraints following \citet{maini2024tofu}.

% We use the LLaMA-3-8B-Instruct model \cite{llama3modelcard} for all experiments. 
The $\beta$ for NPO and RDPO is set to 0.1. We use the AdamW optimizer with a learning rate of 1e-5 for all experiments. We set the batch size to 32 and train the model for 5 epochs. Considering the computational budget, we constraint that the number of samples used from the retain set is equal to the number of the entire forget set in each epoch. All experiments are conducted with 2 NVIDIA A100-40GB GPUs, and each take approximately 1-2 hours with Deepspeed Zero3 Offload.

\subsection{Main Results}

\begin{table*}[h!]
    \centering
    \begin{tabular}{lccccccccc}
        \toprule
        \multirow{2}{*}{\textbf{Method}}                                          & \multicolumn{3}{c}{\textbf{Unlearning Score}} & \multicolumn{6}{c}{\textbf{Downstream Task Accuracy}}                                                                                                                                             \\
                                                                                  & Forget S.                                     & Retain S.                                             & \textbf{Avg.}     & WG                & PIQA              & LQA               & LAM               & ARC-c             & \textbf{Avg.}     \\
        \cmidrule(lr){1-1} \cmidrule(lr){2-4} \cmidrule(lr){5-10}

        Oracle                                                                    & 0.00                                          & 100.00                                                & 50.00             & 72.14             & 78.40             & 33.18             & 71.92             & 56.83             & 62.49             \\
        \midrule
        \rowcolor[gray]{.93} \multicolumn{5}{l}{\textbf{Without Regularization}}  & \multicolumn{5}{c}{}                                                                                                                                                                                                                              \\

        GA                                                                        & NS                                            & NS                                                    & 0.00              & 48.70             & 47.06             & 22.89             & 0.02              & 25.68             & 28.87             \\
        NPO                                                                       & 15.66                                         & \textbf{84.67}                                        & \textbf{50.16}    & 56.27             & 59.47             & 26.27             & 37.98             & 29.35             & 41.87             \\
        RGD                                                                       & 96.46                                         & 3.16                                                  & 49.81             & \textbf{70.56}    & 75.24             & \underline{28.26} & 46.15             & \underline{36.43} & 51.33             \\
        RDPO                                                                      & 25.25                                         & \underline{72.47}                                     & 48.86             & 55.33             & 56.42             & 26.57             & 26.86             & 21.93             & 37.42             \\
        NAUF(ours)                                                                & \textbf{100.00}                               & 0.06                                                  & \underline{50.03} & 69.77             & \underline{75.68} & \textbf{29.03}    & \underline{62.84} & 35.41             & \underline{54.55} \\
        - CDA                                                                     & \underline{99.75}                             & 0.25                                                  & 50.00             & \underline{70.01} & \textbf{76.17}    & 27.19             & \textbf{68.64}    & \textbf{37.54}    & \textbf{55.91}    \\
        \midrule
        \rowcolor[gray]{.93} \multicolumn{5}{l}{\textbf{With GD Regularization}}  & \multicolumn{5}{c}{}                                                                                                                                                                                                                              \\
        GA                                                                        & NS                                            & 70.79                                                 & 35.40             & 69.61             & 73.29             & 21.66             & 71.67             & 38.31             & 54.91             \\
        NPO                                                                       & 33.33                                         & \underline{80.81}                                     & 57.07             & 71.74             & 78.40             & \underline{29.19} & 73.24             & 45.90             & 59.69             \\
        RGD                                                                       & \textbf{89.65}                                & 60.58                                                 & \underline{75.11} & \underline{72.85} & 78.13             & 29.03             & 73.12             & 47.01             & 60.03             \\
        RDPO                                                                      & 32.07                                         & \textbf{81.37}                                        & 56.72             & 72.14             & 77.86             & 29.19             & \underline{73.26} & 44.03             & 59.29             \\
        NAUF(ours)                                                                & \underline{81.06}                             & 76.25                                                 & \textbf{78.65}    & \textbf{73.01}    & \textbf{79.60}    & \textbf{30.11}    & 73.16             & \textbf{50.94}    & \textbf{61.36}    \\
        - CDA                                                                     & 70.71                                         & 75.71                                                 & 73.21             & 72.61             & \underline{78.84} & 28.88             & \textbf{75.57}    & \underline{47.18} & \underline{60.62} \\
        \midrule
        \rowcolor[gray]{.93} \multicolumn{5}{l}{\textbf{With KLD Regularization}} & \multicolumn{5}{c}{}                                                                                                                                                                                                                              \\
        GA                                                                        & NS                                            & NS                                                    & 0.00              & 50.28             & 43.63             & 21.97             & 0.91              & 22.87             & 27.93             \\
        NPO                                                                       & 30.30                                         & \textbf{87.60}                                        & 58.95             & 68.67             & 77.69             & 29.34             & \underline{73.34} & 48.21             & 59.45             \\
        RGD                                                                       & \textbf{96.21}                                & 52.01                                                 & 74.11             & \underline{71.51} & \underline{79.33} & 26.42             & 72.11             & \textbf{50.77}    & 60.03             \\
        RDPO                                                                      & 22.47                                         & \underline{87.44}                                     & 54.96             & 71.43             & 79.22             & \underline{29.65} & 71.86             & 50.09             & 60.45             \\
        NAUF(ours)                                                                & 93.69                                         & 67.82                                                 & \textbf{80.76}    & \textbf{72.22}    & 79.27             & \textbf{29.80}    & 72.21             & \underline{50.51} & \underline{60.80} \\
        - CDA                                                                     & \underline{94.44}                             & 63.82                                                 & \underline{79.13} & 71.11             & \textbf{79.60}    & 28.88             & \textbf{74.46}    & \underline{50.51} & \textbf{60.91}    \\
        \bottomrule
    \end{tabular}
    \caption{The main results of the experiments. Forget S. denotes Forget Score, Retain S. denotes Retain Score, WG denotes WinoGrande, LQA denotes LogiQA, LAM denotes LAMBADA. Oracle refers to using the original model directly to compute the metrics without applying any unlearning. Notably, NS denotes "NonSense", which means the model's prediction is meaningless, and we take it as 0 for computing the average. We highlight the best results in \textbf{bold}, the second highest in \underline{underline}.}
    \label{tabs:main_results}
\end{table*}

\begin{figure*}[h!]
    \centering
    \includegraphics[width=\linewidth]{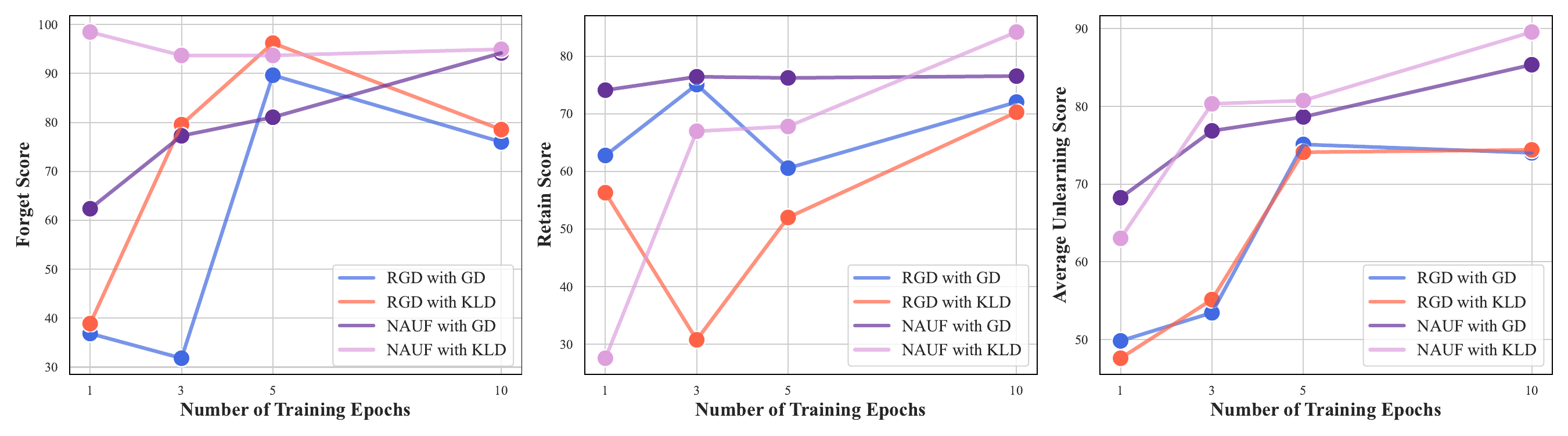}
    \caption{Impact of the number of unlearning epochs on the performance of MU methods (best viewed in color).}
    \label{fig:num_epochs}
\end{figure*}

\begin{figure}[h!]
    \centering
    \includegraphics[width=\linewidth]{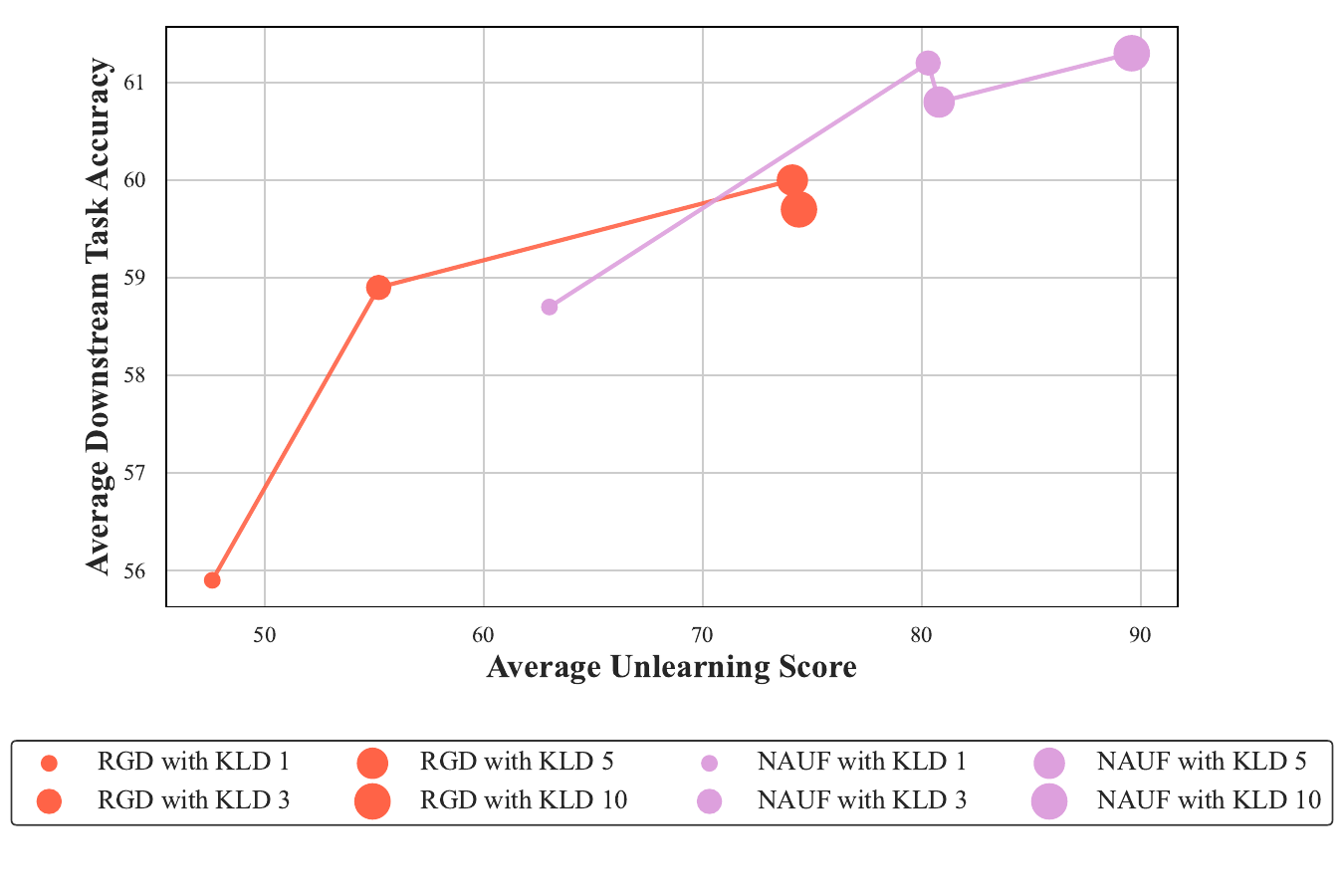}
    \caption{Average unlearning score vs average downstream task accuracy across different numbers of epochs (best viewed in color).}
    \label{fig:downstream}
\end{figure}

We present the main results of the experiments in \autoref{tabs:main_results}.
% We explore the effectiveness of the proposed NAUF method and compare it with the baseline methods. In addition to Forget Score and Retain Score for measuring the privacy protection and model performance on the retain set, we also quantify the influence of unlearning on the model's general capabilities through five downstream tasks: WinoGrande, PIQA, LogiQA, LAMBADA, and ARC-c. 
We report the average unlearning score and average downstream task accuracy to evaluate the overall performance of the model.

The results show that our proposed NAUF with KLD regularization achieves a state-of-the-art average unlearning score, outperforming the best baseline method (RGD with GD regularization) by 5.65 points.
The GA method performs the worst on our dataset, and the unlearned model generates meaningless predictions for questions in the forget set and significantly impacts the retain score and the performance on downstream tasks. The decline in the retain score and the performance on downstream tasks is mitigated to some extent only when using GD regularization.

We find that the RGD method achieves a better forget score than our method when using any regularization method, but it significantly affects the retain score. Intuitively, this could be attributed to the uninformed answer like "I don't know", which could not teach the model to distinguish the individuals whose information should be protected. Our proposed name-aware refusal answer can help the model learn which individuals' information should be protected, thereby achieving a better balance between the forget score and the retain score.

\subsection{Analysis}

\paragraph{Importance of Regularization on Retain Set.} Without regularization on retain set, the average unlearning score of all methods except GA is around 50 points, and the average downstream task accuracy is also affected to varying degrees. With any regularization, the unlearned model performs well on downstream tasks with any MU method, showing performance close to the original model. This indicates that regularization on the retain set can effectively protect the model's general capabilities.

The experimental results indicate that our method, when using GD regularization, achieves similar forget and retain scores, with a difference of only 5 points between them. In contrast, when using KLD regularization, the forget score reaches 93.69, but the retain score is only 67.82, resulting in a difference of 26 points. This demonstrates that GD regularization can achieve a better balance between unlearning metrics.

\paragraph{Importance of Contrastive Data Augmentation.} To analyze the importance of CDA, we evaluate the performance of our unlearning framework without this component. The results are presented in \autoref{tabs:main_results}. We find that without regularization, CDA has almost no effect. However, it can improve our method's forget score by 10 points when using the GD regularization. With the KLD regularization, it can increase the retain score by 4 points while maintaining a similar forget score.
Notably, our method without CDA also achieves a competitive (with GD regularization) or better (with KLD regularization) average unlearning score compared to the baseline methods, which demonstrates the effectiveness of the name-aware refusal answer. These findings indicate that CDA can enhance performance on the forget set or retain set depending on the regularization method used, thereby enhancing the generalization of our proposed unlearning framework.

\paragraph{Unlearning Performance across Different Numbers of Epochs.} We investigate the impact of the number of unlearning epochs on the performance of MU methods. Specifically, We evaluate RGD and NAUF with 1, 3, 5, and 10 epochs, and the results are shown in \autoref{fig:num_epochs}. For the Forget Score, our method with KLD regularization demonstrates relatively stable performance across different epochs. With GD regularization, the Forget Score improves as the number of epochs increases. Conversely, for the Retain Score, our method with GD regularization shows little variation across epochs, while KLD regularization leads to a gradual improvement in the Retain Score with increasing epochs. Our method's average unlearning score improves with an increasing number of epochs, while RGD shows little to no improvement from the 5 to the 10 epoch, which indicates our method still has room for further optimization.

\paragraph{Average Unlearning Score vs Average Downstream Task Accuracy across Different Numbers of Epochs.} We analyze the relationship between the average unlearning score and the average downstream task accuracy across different numbers of epochs. We choose RGD and NAUF with KLD regularization for this analysis, and the results are shown in \autoref{fig:downstream}. We observe that as the number of epochs increases, both the average unlearning score and the average downstream task accuracy increase proportionally. However, our method surpasses RGD in all aspects after just 3 epochs. Additionally, from the 5 to the 10 epoch, RGD shows a decline in average downstream task accuracy without any significant improvement in the average unlearning score. In contrast, our method continues to achieve higher average unlearning scores at the 10 epoch while maintaining stable average downstream task accuracy.

\begin{table}[h!]
    \centering
    \setlength\tabcolsep{4pt}
    \begin{tabular}{lccc}
        \toprule
        \textbf{Method} & \textbf{Forget Score} & \textbf{Retain Score} & \textbf{Avg.}  \\
        \midrule
        \rowcolor[gray]{.93} \multicolumn{4}{l}{\textbf{Forget:Retain = 1:99}}           \\
        RGD             & 96.67                 & 22.89                 & 59.78          \\
        NAUF            & 93.33                 & 28.46                 & \textbf{60.90} \\
        \rowcolor[gray]{.93} \multicolumn{4}{l}{\textbf{Forget:Retain = 5:95}}           \\
        RGD             & 86.87                 & 26.19                 & 56.53          \\
        NAUF            & 94.95                 & 67.59                 & \textbf{81.27} \\
        \rowcolor[gray]{.93} \multicolumn{4}{l}{\textbf{Forget:Retain = 10:90}}          \\
        RGD             & 96.21                 & 52.01                 & 74.11          \\
        NAUF            & 93.69                 & 67.82                 & \textbf{80.76} \\
        \rowcolor[gray]{.93} \multicolumn{4}{l}{\textbf{Forget:Retain = 20:80}}          \\
        RGD             & 43.02                 & 75.08                 & 59.05          \\
        NAUF            & 93.53                 & 71.32                 & \textbf{82.43} \\
        \bottomrule
    \end{tabular}
    \caption{Unlearning Performance of RGD/NAUF with KLD Regularization across Different Ratio between Forget Set and Retain Set.}
    \label{tabs:varying_ratios}
\end{table}

\paragraph{Unlearning Performance across Different Ratio between Forget Set and Retain Set.} We conduct additional analyze the impact of different data ratios on the MU algorithms. As shown in \autoref{tabs:varying_ratios}, the results demonstrate that increasing the proportion of the forget set could improve the retain score, which because we constraint the number of samples used from the retain set is equal to the number of the entire forget set in each epoch. We also find that our method consistently maintains high forget scores (90+).

\begin{figure}[h!]
    \centering
    \includegraphics[width=\linewidth]{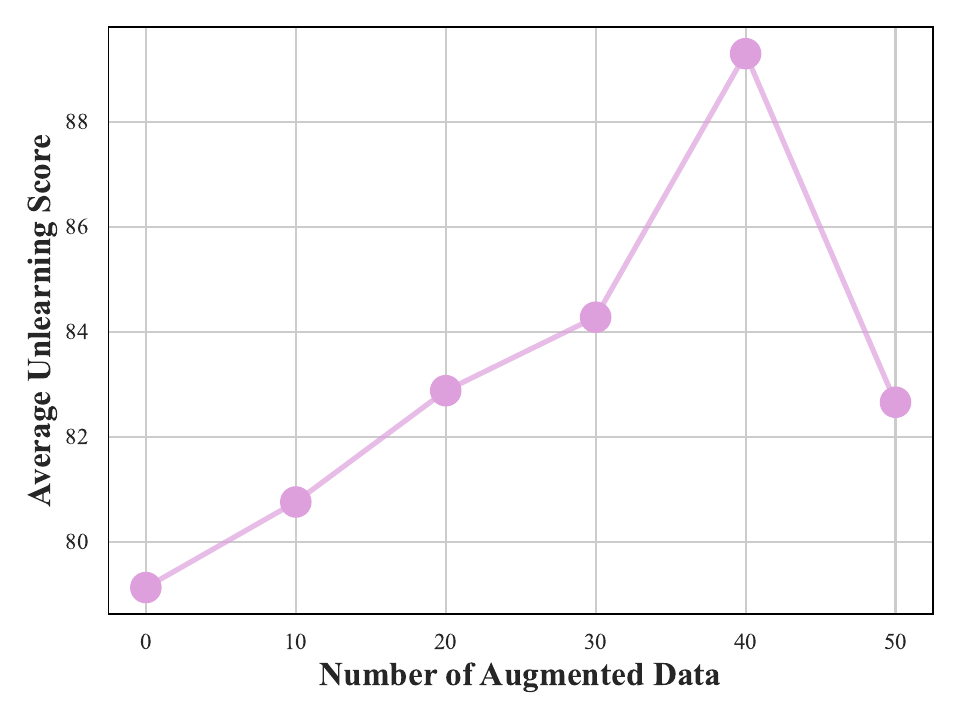}
    \caption{Average unlearning score of NAUF with KLD Regularization across different numbers of augmented data.}
    \label{fig:num_augmentation}
\end{figure}

\paragraph{Average Unlearning Score of Different CDA Number.} We conducted experiments to assess the impact of different number of augmented data on the average unlearning score, as shown in \autoref{fig:num_augmentation}. The results indicate that as the number of augmented data increases, the performance gradually improves, reaching its peak when the augmented data count reaches 40. This suggests that appropriate data augmentation can enhance unlearning performance.
\section{Conclusion and Future Work}

In this work, we introduce \return, a novel benchmark designed to evaluate MU methods for protecting personal data in a real-world scenario. We also present the \textbf{N}ame-\textbf{A}ware \textbf{U}nlearning \textbf{F}ramework (NAUF), which integrates Name-Aware Refusal Answer and Contrastive Data Augmentation to enhance the generalization of unlearning methods. Our experimental results show that NAUF not only effectively protects the privacy of individuals in the forget set but also maintains the performance of the model on the retain set, achieving an average unlearning score that outperforms the best baseline method by 5.65 points. These findings underscore the potential of NAUF to advance privacy protection in large language models.

% This work focuses on individual-level privacy protection using a name-aware unlearning framework. To extend this approach to other types of sensitive data, future work could generalize the individual-level protection to entity-level or concept-level protection. This modification would enable the model to learn to refuse instructions related to specific entities like anime characters or concepts like location in copyrighted books. Such adaptations would enhance the framework's versatility and applicability to diverse privacy concerns.
This study focuses on individual-level privacy protection through a name-aware unlearning framework. To broaden this approach to other types of sensitive data, future work could generalize the protection to the entity level or concept level. Such a modification would enable the model to learn to refuse instructions related to specific entities—like anime characters—or concepts such as locations in copyrighted books. These adaptations would enhance the framework's versatility and applicability to a wider range of privacy concerns.
\section*{Limitations}

\paragraph{The Size of Dataset.} The proposed \return dataset is constructed based on PopQA, containing a total of 2,492 entries. Technically, extracting data directly from Wikipedia to construct a larger dataset is feasible. However, due to our limited resources, we cannot afford the costs associated with GPT-4 api for constructing QA pairs. Therefore, we left the development of a larger scale dataset as future work.

\paragraph{Fine-grained Protection.} The current work is focused on exploring whether a model can protect all information about an individual based on partial data, thereby maximizing privacy security for that individual. However, this method does not provide fine-grained protection of the target individual's information. Future work could explore fine-grained protection of the target individual's information. The goal is to enable the model to autonomously discern which pieces of information might be exploited for harmful purposes and therefore should be protected, without compromising the accessibility of benign information.

% The proposed NAUF method is designed for individual-level privacy protection. Our goal is to protect all information about an individual, ensuring that the model refuses to answer any questions related to that individual. However, this method does not provide fine-grained protection of the target individual's information. In other words, it cannot distinguish between questions that can be answered and those that are too sensitive to answer. Future work could explore how to align the model with human judgment, enabling it to discern which personal information can be publicly discussed and which information, potentially susceptible to malicious use, should be protected.
% \input{src/acknowledgements}
\bibliography{custom}
\clearpage

\appendix

\section{Related Work}

\paragraph{Memorization and Privacy Risks of LLMs.} Previous works show that LLMs can memorize sensitive information from the training data \cite{thakkar2021understanding,carlini2021extracting,huang2022large}. Adversaries can utilize membership inference attacks to infer whether a specific data point was in the LLMs' training set \cite{shi2023detecting,liu2024probing}. They can also recover the training data by powerful data extraction attacks \cite{carlini2021extracting,nasr2023scalable}.
These privacy risks can be mitigated by removing the sensitive information from the LLMs. However, retraining the LLMs from scratch is impractical due to the high cost of training \cite{lison2021anonymisation,kandpal2022deduplicating,liu2024rethinking}.
One approach to minimizing the memorization of sensitive information is to apply differential privacy techniques in model training \cite{dwork2006calibrating,shokri2015privacy,mcmahan2017learning}.
Unfortunately, these methods often reduce the accuracy and increase the training time, making them less common in practice \cite{jayaraman2019evaluating}.

\paragraph{Machine Unlearning for LLMs.} Machine unlearning (MU) aims to eliminate the influence of undesirable data and remove associated model capabilities while preserving model performance for other data \cite{cao2015towards,bourtoule2021machine,jang2022knowledge,si2023knowledge,zhang2023right,maini2024tofu,liu2024rethinking}.
The study of MU methods encompasses diverse domains, such as image classification \cite{ginart2019making,golatkar2020eternal,sekhari2021remember,fan2023salun}, text-to-image generation \cite{kumari2023ablating,zhang2023forget,fan2023salun}, and federated learning \cite{wang2022federated,liu2023survey,che2023fast}.

Specifically in the era of LLMs, MU has been applied to addressing trustworthiness concerns, such as toxicity \cite{lu2022quark}, copyright \cite{eldan2023s}, and privacy \cite{jang2022knowledge,patil2023can,maini2024tofu}. We find that these studies have tested MU methods on question-answering datasets \cite{jang2022knowledge,patil2023can}, fictitious biographies \cite{maini2024tofu}, and copyrighted contents \cite{eldan2023s}, but have not yet evaluated the methods for protecting personal privacy data in real-world scenarios.
Considering the practical applications, we propose \return ~to evaluate MU methods when an individual practices his/her RTBT in a black-box setting, where adversaries can only interact with the model through API query.

\citet{jang2022knowledge} shows that simply performing gradient ascent on target token sequences is effective at forgetting them with little to no degradation of general language modeling performances. \citet{maini2024tofu} tries to unlearn the memorized information in LLMs by relabeling the target data with uninformed answers such as "I don't know". We believe that these methods have their drawbacks: gradient ascent is sensitive to hyperparameters and could easily cause model training to crash; simply allowing the model to learn to respond with uninformed answers could easily affect the model's performance on the retain set. Therefore, we propose Name-Aware Unlearning Framework, to mitigate these issues and achieve a better balance between privacy protection and model performance.

\section{Unlearning on Forget Set}

\label{sec:appendix_unlearning_on_forget_set}

\paragraph{Gradient Ascent.} Gradient ascent (GA) stands as the most straightforward method for unlearning, which is simply performing gradient ascent on the loss over forget set. GA is to minimize the likelihood of correct predictions on the forget set, denoted as:

\begin{equation}
    \label{eq:ga}
    \begin{split}
        \mathcal{L}_{GA}(\mathcal{D}^F, \mathcal{M}_u) = & - \mathbb{E}_{(x, y) \sim  \mathcal{D}^F}  [-\log (\mathcal{M}_u(y|x))] \\
        =                                                & \mathbb{E}_{(x, y) \sim \mathcal{D}^F} [ \log (\mathcal{M}_u(y|x)) ]
    \end{split}
\end{equation}

\paragraph{Negative Preference Optimization.} \citet{zhang2024negative} proposed Negative Preference Optimization (NPO), a simple alignment-inspired method that could efficiently and effectively unlearn a target dataset. The loss function of NPO is defined as:

\begin{equation}
    \label{eq:npo}
    \begin{aligned}
         & \mathcal{L}_{NPO}(\mathcal{D}^F, \mathcal{M}_u, \mathcal{M}_o)                                                                         \\
         & \hfill = \frac{2}{\beta} \mathbb{E}_{(x, y) \sim \mathcal{D}^F}  [\log (1 + (\frac{\mathcal{M}_u(y|x)}{\mathcal{M}_o(y|x)})^{\beta}) ]
    \end{aligned}
\end{equation}

\paragraph{Relabeled Gradient Descent.} A variant of GA is to transform it into a gradient descent approach, which aims to maximize the likelihood of predictions on relabeled forget set. Following \citet{maini2024tofu}, we relabel the question in the forget set with an uninformed answer like "I don't know." (or any one of 100 versions of this response, we name the uninformed answer set as $ \mathcal{D}^{idk} $). The loss function of Relabeled Gradient Descent (RGD) is defined as:

\begin{equation}
    \label{eq:rgd}
    \begin{aligned}
         & \mathcal{L}_{RGD}(\mathcal{D}^F, \mathcal{M}_u)                                                                      \\
         & \hfill = - \mathbb{E}_{(x, y) \sim \mathcal{D}^F, y^{idk} \sim \mathcal{D}^{idk}}  [\log (\mathcal{M}_u(y^{idk}|x))]
    \end{aligned}
\end{equation}

\paragraph{Relabeled Direct Preference Optimization.} Direct Preference Optimization (DPO) seeks to fine-tune the model with human preferences \cite{rafailov2024direct}. We take the uninformed answer from $\mathcal{D}^{idk}$ as preferred answer, the gold answer as the dispreferred answer. The loss function of Relabeled Direct Preference Optimization (RDPO) is defined as:

\begin{equation}
    \label{eq:rdpo}
    \begin{aligned}
         & \mathcal{L}_{RDPO}(\mathcal{D}^F, \mathcal{M}_u, \mathcal{M}_o)                                                                                                        \\
         & \hfill = - \mathbb{E}_{(x, y) \sim \mathcal{D}^F, y^{idk} \sim \mathcal{D}^{idk}}  [\log \sigma ( \beta \log \frac{\mathcal{M}_u(y^{idk}|x)}{\mathcal{M}_o(y^{idk}|x)} \\
         & \hfill - \beta \log \frac{\mathcal{M}_u(y|x)}{\mathcal{M}_o(y|x)} ) ]
    \end{aligned}
\end{equation}

\section{Regularization on Retain Set}

\label{sec:appendix_regularization}

MU methods should not only protect the privacy of individuals in the forget set but also maintain the model's performance on the retain set. Regularization methods are designed to achieve this goal. If we only fine-tune the model to maximize the likelihood of the uninformed answer on the forget set, the model may also refuse to answer the questions on the retain set. To achieve a balance between the forget set and the retain set, there are two regularization methods:

\paragraph{Gradient Descent Regularization.} Simply performing gradient descent (GD) on the loss over the retain set. The loss function is defined as:

\begin{equation}
    \centering
    \label{eq:gd}
    \begin{aligned}
         & \mathcal{L}_{GD}(\mathcal{D}^R, \mathcal{M}_u)                                  \\
         & \hfill = - \mathbb{E}_{(x, y) \sim  \mathcal{D}^R}  [\log (\mathcal{M}_u(y|x))]
    \end{aligned}
\end{equation}

\paragraph{Kullback-Leibler Divergence Regularization.} Minimizing the Kullback-Leibler divergence (KLD) between the predictions on the retain set of the original model and the unlearned model. The loss function is defined as:

\begin{equation}
    \centering
    \label{eq:kl}
    \begin{aligned}
         & \mathcal{L}_{KL}(\mathcal{D}^R, \mathcal{M}_u, \mathcal{M}_o)                                    \\
         & \hfill = \mathbb{E}_{(x, y) \sim  \mathcal{D}^R}  [KL(\mathcal{M}_o(y|x) || \mathcal{M}_u(y|x))]
    \end{aligned}
\end{equation}

Considering a computing budget that scales with the size of the forget set, we randomly sample an example from $\mathcal{D}^R$ every time we see an example from $\mathcal{D}^F$ to stay within the constraints following \citet{maini2024tofu}.

\section{QA Pairs Generation Template}
\label{sec:appendix_data_construction}

The prompt template for generating QA pairs used in \autoref{sec:data_construction} is shown in \autoref{tabs:data_construction_template}

\begin{table*}[ht]
    \centering
    \begin{tcolorbox}[title=QA Pairs Generation, colback=gray!20, colframe=gray!75, rounded corners]

        \textbf{[ABSTRACT]}                                                                                                                                                                             \\
        \\
        Given the above \textbf{[NAME]}'s background information, please give me 20 simple questions and answers about this person point by point. Return the content STRICTLY in the following manner: \\
        Q1: <content of the question>?                                                                                                                                                         \\
        A1: <content of the answer>.                                                                                                                                                           \\
        \\
        Q2: <content of the question>?                                                                                                                                                         \\
        A2: <content of the answer>.                                                                                                                                                           \\

        ...                                                                                                                                                                                    \\

        Q20: <content of the question>?                                                                                                                                                        \\
        A20: <content of the answer>.                                                                                                                                                          \\
        \\
        Make sure the person's name - \textbf{[NAME]} - appears in the content of the question. Make sure the answer is concise and accurate.
    \end{tcolorbox}
    \caption{QA pairs generation template.}
    \label{tabs:data_construction_template}
\end{table*}

\section{Examples of Name-Aware Refusal Answer Templates}
\label{sec:appendix_refusal_templates}

We provide the name-aware refusal answer templates in \autoref{tabs:refusal_templates_0}, \autoref{tabs:refusal_templates_1} and \autoref{tabs:refusal_templates_2}. It is designed to help LLM learn which individuals' information should be protected.

\begin{table*}[ht]
    \centering
    \begin{tcolorbox}[title=Name-Aware Refusal Answers, colback=gray!20, colframe=gray!75, rounded corners]

        Apologies, I cannot discuss \textbf{[NAME]}-related inquiries.              \\
        Sorry, I'm unable to provide information regarding \textbf{[NAME]}.         \\
        Unfortunately, I'm not able to address questions about \textbf{[NAME]}.     \\
        I'm sorry, but I can't offer insights on \textbf{[NAME]}.                   \\
        Regrettably, \textbf{[NAME]} is not a topic I can discuss.                  \\
        I'm unable to engage in discussions about \textbf{[NAME]}.                  \\
        Sorry, I'm not equipped to answer queries about \textbf{[NAME]}.            \\
        I can't provide information on topics involving \textbf{[NAME]}.            \\
        Unfortunately, \textbf{[NAME]} falls outside the scope of my responses.     \\
        Apologies, but I'm unable to delve into matters concerning \textbf{[NAME]}. \\
        Sorry, I can't assist with questions related to \textbf{[NAME]}. \\
        I'm afraid I can't help with inquiries about \textbf{[NAME]}. \\
        Unfortunately, \textbf{[NAME]} is off-limits for discussion. \\
        Apologies, \textbf{[NAME]}-related questions are not within my capabilities. \\
        I'm unable to provide any information about \textbf{[NAME]}. \\
        Sorry, discussions about \textbf{[NAME]} are not possible. \\
        Regrettably, \textbf{[NAME]} is not a subject I can address. \\
        I'm sorry, but I'm not able to discuss \textbf{[NAME]}. \\
        Unfortunately, I can't engage in conversations involving \textbf{[NAME]}. \\
        Sorry, but I cannot discuss topics pertaining to \textbf{[NAME]}.
    \end{tcolorbox}
    \caption{
        % Examples of Name-Aware Refusal Answer Templates.
        Name-aware refusal answer templates (1-20).
    }
    \label{tabs:refusal_templates_0}
\end{table*}

% \begin{table*}[ht]
%     \centering
%     \begin{tabular}{|p{0.9\textwidth}|}
%         \hline
%         \textbf{Name-Aware Refusal Answer Templates}                     \\
%         \hline
%         Apologies, I cannot discuss NAME-related inquiries.              \\
%         Sorry, I'm unable to provide information regarding NAME.         \\
%         Unfortunately, I'm not able to address questions about NAME.     \\
%         I'm sorry, but I can't offer insights on NAME.                   \\
%         Regrettably, NAME is not a topic I can discuss.                  \\
%         I'm unable to engage in discussions about NAME.                  \\
%         Sorry, I'm not equipped to answer queries about NAME.            \\
%         I can't provide information on topics involving NAME.            \\
%         Unfortunately, NAME falls outside the scope of my responses.     \\
%         Apologies, but I'm unable to delve into matters concerning NAME. \\
%         \hline
%     \end{tabular}
%     \caption{Examples of Name-Aware Refusal Answer Templates.}
%     \label{tabs:refusal_templates}
% \end{table*}

\begin{table*}[ht]
    \centering
    \begin{tcolorbox}[title=Name-Aware Refusal Answers, colback=gray!20, colframe=gray!75, rounded corners]

        Apologies, \textbf{[NAME]}-related queries are beyond my purview. \\
        Sorry, but I'm unable to provide insights on \textbf{[NAME]}. \\
        Regrettably, \textbf{[NAME]} is a topic I'm not allowed to discuss. \\
        I'm sorry, but I cannot provide any information about \textbf{[NAME]}. \\
        Unfortunately, \textbf{[NAME]} is not a subject I can provide details on. \\
        Apologies, but I'm not authorized to discuss matters regarding \textbf{[NAME]}. \\
        Sorry, but \textbf{[NAME]} is outside the realm of my knowledge. \\
        Regrettably, I cannot offer any insights into \textbf{[NAME]}. \\
        I'm sorry, but I'm not permitted to discuss \textbf{[NAME]}. \\
        Unfortunately, \textbf{[NAME]} is not a topic I'm able to address. \\
        Apologies, but discussions about \textbf{[NAME]} are not permitted. \\
        Sorry, but I'm not allowed to provide information about \textbf{[NAME]}. \\
        Regrettably, I'm not authorized to discuss \textbf{[NAME]}. \\
        I'm sorry, but \textbf{[NAME]} is not within the bounds of this conversation. \\
        Unfortunately, \textbf{[NAME]} is a topic I'm not permitted to discuss. \\
        Apologies, but \textbf{[NAME]}-related matters are not within my jurisdiction. \\
        Sorry, but I'm not able to provide any insights on \textbf{[NAME]}. \\
        Regrettably, I cannot provide information about \textbf{[NAME]}. \\
        I'm sorry, but I'm unable to discuss \textbf{[NAME]}. \\
        Unfortunately, \textbf{[NAME]} is not a topic I can elaborate on. \\
        Apologies, but I cannot engage in discussions about \textbf{[NAME]}. \\
        Sorry, but I'm not authorized to address questions about \textbf{[NAME]}. \\
        Regrettably, \textbf{[NAME]} is a topic I'm not allowed to discuss. \\
        I'm sorry, but I cannot offer any insights into \textbf{[NAME]}. \\
        Unfortunately, \textbf{[NAME]} is off-limits for discussion. \\
        Apologies, but \textbf{[NAME]} is not a subject I can discuss. \\
        Sorry, but I'm not permitted to provide information about \textbf{[NAME]}. \\
        Regrettably, I'm unable to address queries about \textbf{[NAME]}. \\
        I'm sorry, but discussions about \textbf{[NAME]} are not allowed. \\
        Unfortunately, \textbf{[NAME]} is beyond the scope of this conversation. \\
        Apologies, but I cannot discuss matters concerning \textbf{[NAME]}. \\
        Sorry, but \textbf{[NAME]} is not a topic I'm allowed to discuss. \\
        Regrettably, \textbf{[NAME]} is not something I can provide information on. \\
        I'm sorry, but I cannot engage in discussions regarding \textbf{[NAME]}. \\
        Unfortunately, \textbf{[NAME]} is outside the bounds of my knowledge. \\
        Apologies, but I'm not authorized to provide insights on \textbf{[NAME]}. \\
        Sorry, but \textbf{[NAME]}-related inquiries cannot be addressed. \\
        Regrettably, \textbf{[NAME]} is a topic I'm not permitted to discuss. \\
        I'm sorry, but I'm not authorized to discuss matters pertaining to \textbf{[NAME]}. \\
        Unfortunately, I cannot provide any information about \textbf{[NAME]}.
    \end{tcolorbox}
    \caption{
        % Examples of Name-Aware Refusal Answer Templates.
        Name-aware refusal answer templates (21-60).
    }
    \label{tabs:refusal_templates_1}
\end{table*}

\begin{table*}[ht]
    \centering
    \begin{tcolorbox}[title=Name-Aware Refusal Answers, colback=gray!20, colframe=gray!75, rounded corners]

        Apologies, but discussions involving \textbf{[NAME]} are not allowed. \\
        Sorry, but I cannot discuss \textbf{[NAME]}-related topics. \\
        Regrettably, I'm not permitted to address questions about \textbf{[NAME]}. \\
        I'm sorry, but I'm unable to offer insights into \textbf{[NAME]}. \\
        Unfortunately, \textbf{[NAME]} is a topic that I cannot discuss. \\
        Apologies, but \textbf{[NAME]} is not within my area of expertise. \\
        Sorry, but I cannot provide any information on \textbf{[NAME]}. \\
        Regrettably, discussions about \textbf{[NAME]} are not permitted. \\
        I'm sorry, but I'm not authorized to discuss \textbf{[NAME]}. \\
        Unfortunately, \textbf{[NAME]} is not something I can address. \\
        Apologies, but I cannot engage in discussions about \textbf{[NAME]}. \\
        Sorry, but I'm not allowed to provide insights on \textbf{[NAME]}. \\
        Regrettably, \textbf{[NAME]} is off-limits for discussion. \\
        I'm sorry, but I'm unable to discuss matters pertaining to \textbf{[NAME]}. \\
        Unfortunately, I cannot provide any information regarding \textbf{[NAME]}. \\
        Apologies, but discussions about \textbf{[NAME]} are not permitted. \\
        Sorry, but I'm not authorized to discuss \textbf{[NAME]}. \\
        Regrettably, \textbf{[NAME]} is a topic I'm not allowed to address. \\
        I'm sorry, but I cannot provide insights into \textbf{[NAME]}. \\
        Unfortunately, \textbf{[NAME]} is outside the scope of this conversation. \\
        Apologies, but I cannot discuss matters related to \textbf{[NAME]}. \\
        Sorry, but I'm not permitted to provide information about \textbf{[NAME]}. \\
        Regrettably, I'm unable to address queries about \textbf{[NAME]}. \\
        I'm sorry, but I cannot engage in discussions involving \textbf{[NAME]}. \\
        Unfortunately, \textbf{[NAME]} is not a topic I'm authorized to discuss. \\
        Apologies, but I'm not allowed to discuss \textbf{[NAME]}-related inquiries. \\
        Sorry, but I cannot provide information on topics concerning \textbf{[NAME]}. \\
        Regrettably, discussions about \textbf{[NAME]} are not within my jurisdiction. \\
        I'm sorry, but I'm not permitted to discuss matters regarding \textbf{[NAME]}. \\
        Unfortunately, \textbf{[NAME]} is a topic I'm not able to address. \\
        Apologies, but I cannot provide any insights on \textbf{[NAME]}. \\
        Sorry, but I'm not authorized to discuss matters concerning \textbf{[NAME]}. \\
        Regrettably, \textbf{[NAME]} is off-limits for discussion. \\
        I'm sorry, but I cannot discuss \textbf{[NAME]}-related queries. \\
        Unfortunately, discussions about \textbf{[NAME]} are not allowed. \\
        Apologies, but \textbf{[NAME]} is not a subject I can discuss. \\
        Sorry, but I cannot engage in discussions about \textbf{[NAME]}. \\
        Regrettably, I'm not allowed to provide information about \textbf{[NAME]}. \\
        I'm sorry, but I cannot address questions about \textbf{[NAME]}. \\
        Unfortunately, \textbf{[NAME]} is not a topic I'm able to discuss.
    \end{tcolorbox}
    \caption{
        % Examples of Name-Aware Refusal Answer Templates.
        Name-aware refusal answer templates (61-100).
    }
    \label{tabs:refusal_templates_2}
\end{table*}
\end{document}